\documentclass[11pt, letter, onecolumn]{article}
\usepackage{times}
\usepackage{latexsym}
\usepackage{indentfirst}

\usepackage{booktabs}
\usepackage{url}
\usepackage{listings}
\usepackage{color}
\usepackage{graphicx}
\usepackage{appendix}
\definecolor{dkgreen}{rgb}{0,0.6,0}
\definecolor{gray}{rgb}{0.5,0.5,0.5}
\definecolor{mauve}{rgb}{0.58,0,0.82}
\usepackage{caption}
\usepackage{subcaption}
\usepackage[english]{babel}
\usepackage{blindtext}

\usepackage{times}                                       
\usepackage{amsmath}                                    

\usepackage{amssymb}     

\usepackage[export]{adjustbox}
\usepackage{footnote}

\makeatletter
\let\NAT@parse\undefined
\makeatother
\usepackage{hyperref}  
\usepackage{cleveref}

\usepackage{geometry}
\newgeometry{vmargin={20mm}, hmargin={17mm,20mm}}   

\title{GAN based Data Augmentation to Resolve Class Imbalance}

\author{
  Sairamvinay Vijayaraghavan \\ \and
  Terry Guan \\ \and 
  Jason (Jinxiao) Song
  }

\begin{document}
\maketitle

\section{Introduction}
Online shopping has become a major trend to purchase stuff and credit card is one of the most common pay online. So, it is not surprising to see many potential credit card frauds. In fact, the number of credit card frauds has been growing as technology grows and people can take advantage of it. Therefore, it is very important to implement a robust and effective method to detect such frauds. The machine learning algorithms are appropriate for these tasks since they try to maximize the accuracy of predictions and hence can be relied upon. However, there is an impending flaw where-in machine learning models may not perform well due to the presence of an imbalance across classes distribution within the sample set. So, in many related tasks, the datasets have a very small number of observed fraud cases (sometimes around 1\% positive fraud instances found). Therefore, this imbalance presence may impact any learning model's behavior by predicting all labels as the majority class, hence allowing no scope for generalization in the predictions made by the model. We chose a relevant paper  "Using generative adversarial networks for improving classification effectiveness in credit card fraud detection"\cite{FIORE2019448}, as our primary inspiration for training Generative Adversarial Network(GAN) to generate a large number of convincing (and reliable) synthetic examples of the minority class that can be used to alleviate the class imbalance within the training set. In this project, we present a similar approach to the paper\cite{FIORE2019448} where we focus on a small subset of the provided dataset to address the class imbalance issue and generate enough synthetic positive samples by training GANs and we compare with the existing vanilla oversampling method for a cohesive comparison. 

\section{Motivation}
Credit card fraud loses consumers and banks billions of dollars yearly. It is important that companies are able to detect credit card frauds so that it protects banks' funds. More importantly, we ensure that customers don't pay for items that they did not purchase. In order to solve this problem, we have to ensure that the learning models are able to generalize learning from complex data patterns across the datasets and effectively reduce the false positives within the fraud detection system. However, since non-fraudulent transactions happen much more often than fraudulent transactions in real, the dataset can be highly class-imbalanced. In order to train smarter learning algorithms for generalizing patterns amidst such imbalance, we decided to leverage different types of data augmentation methods before training classifiers for improving the effectiveness of the learning models.

Since there is such a large class imbalance, it can impede the predictive performance of our learning classification models. A popular method for dealing with this class imbalance is to oversample the minority class or undersample the majority class. A simple method for undersampling is simply to delete random data from the majority class and ensure matching of classes distribution. However, this can cause important data to be deleted and also drastically reduce the sample size of our training data. Moreover, deleting these samples would possibly change the decision boundary of the model while learning the classifiers. Vanilla oversampling involves duplication of random samples from the minority training class for increasing the number of samples within the minority class to improve the training. This increases the sample size by ensuring an equal number of instances in both the classes. However, this can cause overfitting in certain models because of the lack of adding new information about the minority class and hence a lack of generalization.

Following in-line with this work\cite{FIORE2019448}, our team chose to implement generative adversarial networks(GANs) to generate and combine synthetic minority class samples to the already existing minority class samples and then compare the performance to that of vanilla oversampling and also raw data (without any data augmentation). Generative adversarial networks are a framework of neural networks in which two models: generator and discriminator are competent in beating each other's learning scheme. The discriminator model learns to determine whether a sample is generated from the model (fake) or belonging to the actual data distribution (real). Similarly, "the generative model can be thought of as analogous to a team of counterfeiters, trying to produce fake currency and use it without detection, while the discriminator model is analogous to the police, trying to detect the counterfeit currency. Competition in this game drives both teams to improve their methods until the counterfeits are indistinguishable from the genuine articles."\cite{goodfellow2014generative}

\section{Data}
The dataset\cite{Pozzolo} obtained from Kaggle that we are using just as the paper did \cite{FIORE2019448} contains transactions made by European cardholders in September of 2013. All transactions occurred in two days, where it had 284,807 transactions and 492 of those transactions are considered fraud. It is high class imbalanced with only 0.172\% of transactions being fraudulent. The dataset contains 30 features, only 2 of which are not PCA transformations, Time, and Amount, which was retained from the real raw data collections.
\par Of the 284,807 transactions we chose to work with a reduced sample size for this project. We decided to work on a training dataset of 10,000 samples and then test 5,000 held-out samples for our classification experiments. We split based on the positive samples such that the training samples have 315 (67\%) positive values and the test set to have 158 (33\%) positive samples. We then picked random negative samples and added them to our train and test sets so that we ensured the training size was 10000 and the test size was 5000. We reduced the dataset due to training time limitations and for the simplicity of this project's scope. Since the dataset was having a heavy imbalance towards negative (non-fraudulent) samples, we decided to retain all the positive (fraudulent) data points. We removed the duplicates within our dataset and split our dataset into train and test sets.  We also decided to normalize our feature values by applying two scaling methods: standard scaling using mean and variance, and before training classifiers, we then applied min-max scaling. These methods ensured our features have a Gaussian-like distribution and all the feature values are within 0 and 1 ranges.
\par Since the data is highly imbalanced, data augmentation on the minority class can be used to improve the classification results. Even after reducing the dataset size, only ~3\% of the training data belongs to the positive class. To deal with the class imbalance, we propose oversampling the minority class via two methods: random oversampling and GAN-based data augmentation by adding as many positive samples such that there are equal number of negative and positive samples in the dataset before training the classifiers. Then after inducing fake samples, we compare the results of the two by training the same set of classifiers.

\section{Approach}
In the beginning, not only did we just implement the paper, but we also needed to understand how GANs work particularly on the mechanics of GAN such as the relationship between generator and discriminator before training and how it can help address the class imbalance problem. Then, we designed our experiments by following very closely what the paper did wherein they compare data augmentation with adversarial training with existing oversampling methods before classifying the models. Our focus was to validate the significance of adversarial training in data generalization while training classifiers while also smartly alleviating class imbalance. Self-supervised adversarial training before classification helps in inducing fake examples in order to resolve the class imbalance. We choose to train Generative adversarial networks (GAN) for our project just as followed by the paper for augmenting our dataset before training. Then we trained a simple binary classifier for resolving the main problem of fraud detection. For the binary classifier, we worked with traditional classifiers such as logistic regression, multi-layer perceptron, Support Vector Machines (SVM), and decision trees. We compare the effectiveness of data augmentation, we even chose to include vanilla oversampling for our comparisons. \\
The framework listed on the paper we are following is:
\begin{description}
\item 1. Isolate all fraudulent (positive) examples in the training set T. Denote the resulting set by $P$ of size 315 samples.
\item 2. Use set $P$ as a training set for a GAN, tuning its hyper-parameters.
\item 3. Use the trained generator $G$, of the GAN to generate synthetic examples $P^\prime$, receiving as input random noise.
\item 4. Merge $P^\prime$ into the existing training set $T$ to get the final training set $T^{augm}$. Then train our classifier $C$ on $T^{augm}$ and compare the results with two other cases: with oversampling applied in place of generalized data augmentation and without any oversampling techniques into the dataset, training it raw.

\end{description}

\par Our project replicated the paper by using all the 30 features of the dataset. The GAN architecture we followed was almost identical to the paper. We had used 2 hidden layers for the generator both with ReLU (Rectified Linear Unit) activation and the output layer having a sigmoid activation function. The first hidden layer has 100 neuron units, and the second hidden layer has 30 neuron units. The generator takes in an input noise vector of 100 dimensions and the output layer provides a matching vector to any existing sample with a dimensionality of 30, equalling the number of features trained. For the discriminator, we used 3 hidden layers, each of which has a sigmoid activation and 36 neuron units. The discriminator took in an existing potential input vector of size 30 and it predicts whether it was generated or real. So, our output layer was a single perceptron unit with sigmoid activation.
\par We implemented slight changes to the existing paper architecture. We included a batch normalization layer applied before the output layer in the generator model which stabilized the learning of the generator and to converge quicker. We also added a dropout layer with a rate of 20\% with binary cross-entropy for our loss function in order to reduce overfitting within the discriminator. The paper used a learning rate from the set of potential values: $5\times10^{-4}$, $5\times10^{-3}$, and $5\times10^{-2}$. We started with a learning rate of $5\times10^{-2}$, then $5\times10^{-3}$, and $5\times10^{-4}$, but found that all those learning rates did not converge at all and as a result, the discriminator never learned. By trial, we applied a learning rate of $1\times10^{-5}$ which worked the most optimal for our architecture. Since the paper did not explicitly mention how many epochs they trained with, we found out empirically that 10000 epochs (indicating a sufficient time for learning GANs) worked best for our project.
\par After training GANs, we ensured that we generated enough positive samples so that our training set has a 50-50 \% even split of both the class-based samples. So, we generated 9370 additional positive samples and merged them with the original training set in order to ensure that 9685 samples were present in each of the classes, and hence our training set size was increased to 19370 for each of the oversampling methods. We then trained more classifiers than what the paper actually followed. We additionally trained SVM, decision trees, logistic regression, and multi-level perceptron(MLP) models. The paper has implemented the GAN along with a classifier as one single combined model, however, we chose to dissociate the classifier from the GANs so that we can train the classifiers exclusively from data generation tasks. We implemented the same architecture for the MLP as the paper. We used two hidden layers each with 30 hidden units, with the first hidden layer having a ReLU activation while the second layer had a sigmoid activation. The output layer has a softmax activation for predicting 2 probabilities for each of the classes (positive and negative). We used categorical cross-entropy with Adam optimizer and the same learning rate $1\times10^{-5}$. Finally, we compared the predictive performance on 3 different training sets: the one with augmented generated data from the GANs, the one with vanilla duplicated oversampling, and the one with just the raw data.

\section{Results}
\begin{figure}[h!]
\center
\begin{subfigure}{.5\textwidth}
  \centering
  \includegraphics[width=\linewidth]{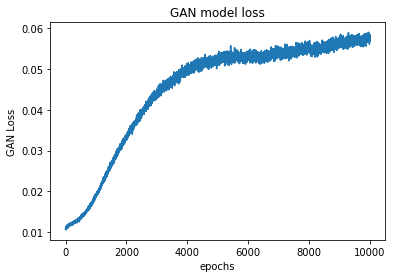}
  \caption{GAN loss}
  \label{fig:sub1}
\end{subfigure}%
\begin{subfigure}{.5\textwidth}
  \centering
  \includegraphics[width=\linewidth]{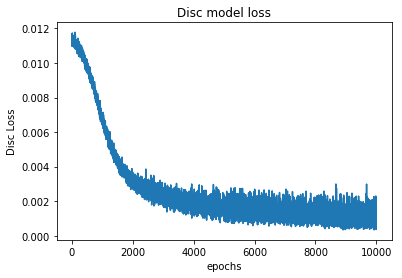}
  \caption{Discriminator loss}
  \label{fig:sub2}
\end{subfigure}
\caption{GAN vs Discriminator loss }
\label{fig:test_gan}
\end{figure}

\begin{figure}[h!]
\center
\begin{subfigure}{.5\textwidth}
  \centering
  \includegraphics[width=\linewidth]{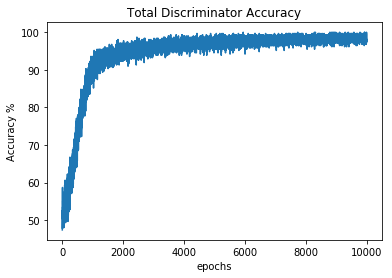}
  
  \label{fig:sub1}
\end{subfigure}%
\caption{Discriminator Accuracy}
\label{fig:test_disc_acc}
\end{figure}

We have observed some unexpected and unusual scenarios which happened during the development phase. For example, the loss of the GAN was rising as the discriminator kept learning (Figures \ref{fig:test_gan},\ref{fig:test_disc_acc}) and we found this as a correct step. This is because the increase in the loss of the generator implied a stronger generation of fake samples which are totally deviant from the real existent data and hence the generator was maximizing its loss while the discriminator was learning better (with a reducing loss and increasing accuracy) in identifying the prediction across real and fake samples.

Table \ref{tab:table1} and Table \ref{tab:table3} are a direct comparison between vanilla oversampling and GAN(Data Augmentation). Based on the table, the implementation of using GAN has better metric scores in oversampling in all classifier models. The most important difference is the precision and F1 score in SVM and logistic regression also shows the performance improvement with the implementation of using GAN. 

\begin{figure}[h!]
\center
\begin{subfigure}{.4\textwidth}
  \centering
  \includegraphics[width=\linewidth]{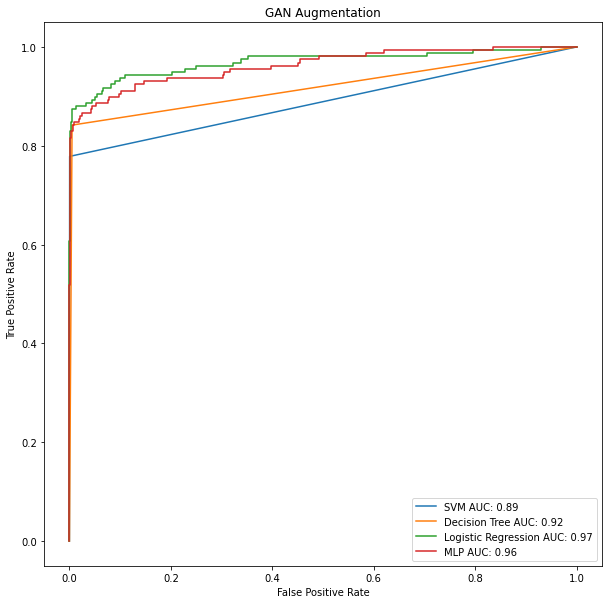}
  \caption{Data Augmentation}
  \label{fig:sub1}
\end{subfigure}%
\begin{subfigure}{.4\textwidth}
  \centering
  \includegraphics[width=\linewidth]{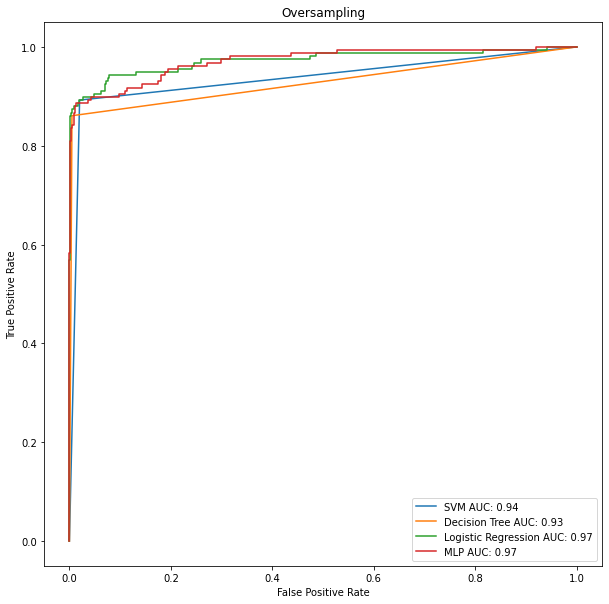}
  \caption{Over-Sampling}
  \label{fig:sub2}
\end{subfigure}
\caption{AUC Comparision between Data Aug VS Oversampling}
\label{fig:test_auc}
\end{figure}

In Figure \ref{fig:test_auc}, it is a direct comparison of the receiver operatic curves along with the area under the curve for each classifier when trained across both the GAN-based vs vanilla oversampling-based data augmentation applied in the training set. In comparison with the F1 scores across tables \ref{tab:table1} and \ref{tab:table3}, the GAN-based augmentation does better in comparison to the vanilla oversampling for all the classifiers but decision trees. Similarly, GAN augmentation performs on par and/or better than the other experiments because it helps generalization more than the other methods in the training. All metric scores are similar across each of the 3 tables besides MLP (which is the best for the GAN augmentation). Apparently, for some simple classifiers, original raw data (table \ref{tab:table2}), actually has slightly better metric scores such as precision and F1 than vanilla oversampling. We ensured that GAN augmentation has a comparable and better effect in generalized learning with a slightly better result than vanilla oversampling and very much on par results with raw training on the data without any augmentation. 

\section{Future Work}
We seek an extension by implementing a statistically better method that involves duplicating samples by creating synthetic examples for the minority class using the data distribution of the original data. The most popular method for synthesizing new examples through this method is Synthetic Minority Oversampling Technique(SMOTE)\cite{2002}. It creates examples that are close in feature space and creates a point with some noise associated with that point. There can be more experiments to compare this oversampling method along with our existing experiments compared to random oversampling and GAN oversampling. Additionally we could have incorporated a hybrid augmentation technique involving both SMOTE and GAN based Augmentation to improve our sample size \cite{10.1007/978-3-030-36945-3_7}.\\

Similarly, we provide another possible extension of applying a different GAN extension framework such as Wassterstein GANs\cite{ENGELMANN2021114582} that could possibly provide more quality in producing the data augmented samples for improving our training results. We could also train a more complex deep learning architecture that learns the classifier along with the generator and discriminator as a three-player adversarial game, and hence generates real-like minority class data which appears very close to the original data in a geometric space \cite{mullick2019generative}.

\section{Conclusion}

Overall, with the basis of inspiration from the paper \cite{FIORE2019448}, we have presented an alternative approach to address the class imbalance dataset problem by not just following a plain unintelligent oversampling, but also allowing a more generalizable data augmentation method. We have demonstrated that not only GAN is a good approach to address the class imbalance problems but it can be a better approach than normal random oversampling by focusing on a more generalizable sample generation which allows more intelligently induced synthetic samples. Hence, we can see that this data augmentation approach is much more practical and can be used in many applications that involve class imbalance which can be effectively resolved with synthetic examples generated by GAN so as to improve predictive performance. 

\section{Contribution}
Terry Guan, Jason (Jinxiao) Song and Sairamvinay Vijayaraghavan have worked together and participated evenly on the entire project in all aspects: idea developments, coding, and report writing.
\newpage
\section{Supplemental tables and graphs}
\begin{table}[h!]
  \begin{center}
    \caption{Data Augmentation}
    \label{tab:table1}
    \begin{tabular}{l|l|r|l|l|l|l|l} 
      \textbf{Augmentation} & \textbf{Model} & \textbf{Accuracy \%} & \textbf{Recall/Sensitivity}& \textbf{Precision}& \textbf{F1}& \textbf{Spec}& \textbf{AUCROC}\\
      \hline
      GAN &SVM	&99.26&0.79&0.98&0.87&0.78&0.89\\
      \hline
      GAN &DT &99.08	&0.84	&0.84	&0.84	&0.84	&0.92\\
      \hline
      GAN &LOGReg &99.0	&0.72	&0.97	&0.83&0.72	&0.97\\
      \hline
      GAN &MLP	&99.08	&0.73	&0.97	&0.83&0.73	&0.96\\
      \hline
    \end{tabular}
  \end{center}
\end{table} 

\begin{table}[h!]
  \begin{center}
    \caption{OverSampling}
    \label{tab:table3}
    \begin{tabular}{l|l|r|l|l|l|l|l} 
      \textbf{Augmentation} & \textbf{Model} & \textbf{Accuracy \%} & \textbf{Recall/Sensitivity}& \textbf{Precision}& \textbf{F1}& \textbf{Spec}& \textbf{AUCROC}\\
      \hline
      OverSamp &SVM	&97.72&0.89&0.59&0.71&0.89&0.94\\
      \hline
      OverSamp &DT &99	&0.86	&0.85	&0.86	&0.86	&0.93\\
      \hline
      OverSamp &LOGReg &97.86	&0.89	&0.61	&0.72 &0.89	&0.97\\
      \hline
      OverSamp &MLP	&98.72	&0.84	&0.77	&0.81&0.84	&0.97\\
      \hline
    \end{tabular}
  \end{center}
\end{table} 
\begin{table}[h!]
  \begin{center}
    \caption{Raw Data}
    \label{tab:table2}
    \begin{tabular}{l|l|r|l|l|l|l|l} 
      \textbf{Augmentation} & \textbf{Model} & \textbf{Accuracy \%} & \textbf{Recall/Sensitivity}& \textbf{Precision}& \textbf{F1}& \textbf{Spec}& \textbf{AUCROC}\\
      \hline
      Raw (None)&SVM	&99.32&0.79&0.99&0.88&0.79&0.9\\
      \hline
      Raw (None) &DT &99.04	&0.85	&0.85	&0.85	&0.85	&0.92\\
      \hline
      Raw (None) &LOGReg &99.14	&0.73	&1	&0.84&0.73	&0.97\\
      \hline
      Raw (None) &MLP	&97.6 &0.24	&1	&0.39&0.24	&0.95\\
      \hline
    \end{tabular}
  \end{center}
\end{table}


\clearpage
\newpage
\bibliographystyle{plain}
\newpage
\nocite{*}

\bibliography{bibliography.bib}
\end{document}